\documentclass[%
 preprint,
 amsmath,
 amssymb,
 aps,
 prr,
 longbibliography,
]{revtex4-1}

\usepackage{algorithm}
\usepackage{algpseudocode}
\usepackage{algorithmicx}

\usepackage{CJK}
\usepackage{amsmath}
\usepackage{amssymb}
\usepackage{bbold}
\usepackage{color}
\usepackage{ulem}
\usepackage{braket}
\usepackage{here}
\usepackage{graphicx}
\usepackage{hyperref}
\usepackage{dcolumn}
\usepackage{bm}
\usepackage{natbib}
\usepackage{multirow}

\newcommand{\form}{$\mathrm{H}_2\mathrm{CO}\;$}
\newcommand{\kayser}{$\mathrm{cm}^{-1}\;$}
\newcommand{\magenta}[1]{{#1}}  

\begin{document}
\begin{CJK*}{GB}{} 
\title{Neural Network Matrix Product Operator: \\A Multi-Dimensionally Integrable Machine Learning Potential}

\author{Kentaro Hino}
\email{hino@theoc.kuchem.kyoto-u.ac.jp}
\affiliation{Department of Chemistry, Graduate School of Science, Kyoto University, Kitashirakawa Oiwake-cho, Sakyo-ku Kyoto, 606-8502, Japan}

\author{Yuki Kurashige}
\email{kura@kuchem.kyoto-u.ac.jp}
\affiliation{Department of Chemistry, Graduate School of Science, Kyoto University, Kitashirakawa Oiwake-cho, Sakyo-ku Kyoto, 606-8502, Japan}
\affiliation{FOREST, JST, Honcho 4-1-8, Kawaguchi, Saitama 332-0012, Japan}
\affiliation{CREST, JST, Honcho 4-1-8, Kawaguchi, Saitama 332-0012, Japan}
\date{\today}
\begin{abstract}
A neural network-based machine learning potential energy surface (PES) expressed in a matrix product operator (NN-MPO) is proposed.
The MPO form enables efficient evaluation of high-dimensional
integrals that arise in solving the time-dependent and
time-independent Schr\"odinger equation, and effectively
overcomes the so-called curse of dimensionality.
This starkly contrasts with other neural network-based machine learning PES methods, such as multi-layer perceptrons (MLPs), where evaluating high-dimensional integrals is not straightforward due to the fully connected topology in their backbone architecture.
Nevertheless, the NN-MPO retains the high representational capacity of neural networks.
NN-MPO can achieve spectroscopic accuracy with a test mean absolute error (MAE) of 
3.03 \kayser for a fully coupled six-dimensional $ab\;initio$ PES,
using only 625 training points distributed across a 0 to 17,000 \kayser energy range.
Our Python implementation is available at
\url{https://github.com/KenHino/Pompon}.
\end{abstract}

\maketitle
\end{CJK*}

\section{Introduction}
The developments of machine learning potential energy surfaces (PESs) in quantum chemistry and condensed matter physics have been accelerating alongside the advancements in a wide range of computational technologies
\cite{behler2015behler,behler2016behler,schutt2017muller,deringer2019csanyi,takamoto2022ibuka}.
Neural network potentials, in particular, have garnered significant attention for their ability to reproduce the accurate but expensive $ab\;initio$ electronic structure calculations.
For universal interatomic potentials, the model input is often replaced by user-defined descriptors that are augmented with atomic numbers and symmetry-consistent features, such as interatomic distances
\cite{behler2007parrinello,behler2011behler,unke2019meuwly,batzner2022kozinsky,batatia2022csanyi,takamoto2022li}.
The incorporation of atomic descriptors enables the model to estimate individual atomic energies. Consequently, by summating these atomic contributions, the model can predict the total potential energy of systems comprising arbitrary numbers of atoms.

Neural network potentials commonly employ multi-layer perceptrons (MLPs) as their foundational architecture.
These structures are characterized by their extensive parameter spaces, which allow them to approximate highly complex functions.
Notably, despite the substantial number of parameters involved, recent research has demonstrated that overparameterized neural networks, when trained using stochastic optimization techniques, exhibit the ability to mitigate complexity-induced errors and achieve robust generalization performance
\cite{zhang2021vinyals,frankle2019carbin,allen-zhu2019song}.
Trained neural network potentials can predict properties without computationally expensive $ab\;initio$ calculations.
In particular, using neural network potential as a force field for classical molecular dynamics simulation that treats the nuclear motions as classical particles is becoming a promising approach \cite{zhang2018e,wang2018e}.
On the other hand, in the context of quantum many-body simulations,
if we consider replacing the potential $\hat{V}$
of Hamiltonians with a machine learning potential $V(\mathbf{x})$,
\begin{equation}
  \label{eq:schrodinger}
  \hat{H} = \hat{T} + \hat{V}
  = -\sum_{i=1}^{n}\frac{\hbar^2}{2M_i}\nabla_{i}^2 + V(\mathbf{x}),
\end{equation}
one of the bottlenecks lies in the evaluation of
the multi-dimensional integrals
between wavefunction $\Psi$ and Hamiltonian $\hat{H}$,
which appears not only in the evaluation of the expectation value
$\Braket{\Psi|\hat{H}|\Psi}$
but also in the updating procedures of variational parameters
based on the variational principle
$\Braket{\delta\Psi|\hat{H}| \Psi}=0$,
which is repeatedly called in both time-dependent and time-independent problems.
Because of the fully connected topology of MLPs,
they cannot efficiently evaluate multi-dimensional integrals
and often resort to an approximated method such as Monte Carlo integration.
Therefore, it is not efficient for nuclear wavepacket simulation
such as multi-dimensional time-dependent Hartree (MCTDH) \cite{beck2000meyer,meyer1990cederbaum}, and instead,
the sum-of-products (SOP) forms \cite{jackle1996meyer,panades-barrueta2020pelaez,schroder2020schroder,song2022meng,koch2019burghardt,manzhos2006carrington, nadoveza2023pelaez, sasmal2024vendrell}
\begin{equation}
  \label{eq:sop}
  V_{\text{SOP}}(\mathbf{x}) = \sum_{\boldsymbol{\rho}} c_{\boldsymbol{\rho}}\prod_{i}^{n} v_{\rho_i}(x_i)
\end{equation}
like polynomials or high-dimensional model representations (HDMR) \cite{li2001rabitz, manzhos2006carrington-2},
which approximate an $n$-body function as a sum of up to $k$-body functions,
\begin{equation}
  \label{eq:hdmr}
  V_{\text{HDMR}}(\mathbf{x}) =
  \sum_{i_1}^{n} c_{i_1} v_{i_1}(x_{i_1}) + \sum_{{i_1}<{i_2}}^{n} c_{{i_1}{i_2}} v_{{i_1}{i_2}}(x_{i_1}, x_{i_2}) + \cdots
\end{equation}
have been employed
because both can evaluate multi-dimensional integrals between
wavefunction $\Ket{\sigma_1}\Ket{\sigma_2}\cdots\Ket{\sigma_f}$
by the sum of products of one-dimensional integrals
$\Braket{\sigma_i^\prime|v_{\rho_i}|\sigma_i}$
 and the sum of $k$-dimensional integrals
$\Braket{\sigma_{i_1}^\prime\sigma_{i_2}^\prime\cdots\sigma_{i_k}^\prime|v_{i_1i_2\cdots i_k}|\sigma_{i_1}\sigma_{i_2}\cdots\sigma_{i_k}}$,
respectively.

Here, we present a machine learning potential facilitated by tensor
network (TN), which offers both high representational power as MLPs
and efficient evaluation of multi-dimensional integrals as SOPs and
HDMRs.
TNs have been initially developed for quantum many-body problems.
In particular, the density matrix renormalization group (DMRG)
\cite{white1992white,white1993white}
is the most well-known ansatz,
established as a method approaching exact solutions
in various quantum many-body systems,
such as spin systems \cite{white1992white,white1993white},
electronic systems \cite{chan2002head-gordon,chan2011sharma},
and phonon systems \cite{baiardi2017reiher}.
The matrix product state (MPS) is known as a
representation of the many-body wavefunction
in a one-dimensional TN called tensor train (TT).
This formalism efficiently expresses the exponentially large tensor
$A_{i_1i_2\cdots i_n} \in {\mathbb{K}^{d_1\times d_2\times\cdots\times d_n}}$
as a one-dimensional contraction of polynomially-sized three-rank tensors
$\{G^{[i]}_{\alpha_{i-1}i\alpha_i} \in {\mathbb{K}^{D_{i-1}\times d_i\times D_i}}\}$;
$$
A_{i_1i_2\cdots i_n}
\approx
\sum_{\alpha_1\alpha_2\cdots\alpha_{n-1}} G^{[1]}_{i_1\alpha_1}G^{[2]}_{\alpha_1i_2\alpha_2}G^{[3]}_{\alpha_2i_3\alpha_3}\cdots G^{[n]}_{\alpha_{n-1}i_n}.
$$
Because TNs can reproduce the original exponentially large 
tensor product coefficients $A_{i_1i_2\cdots i_n}$ in the limit of
$\{D_i\}$, $i.\;e.$ rank of connecting indices $\{\alpha_i\}$, 
TNs should be able to represent any function for machine learning tasks with the desired accuracy.
Furthermore, TNs can extract important features from multi-dimensional data through low-rank approximations, which is key to reducing parameters and preventing overfitting in machine learning.
$Schwab$ and $Stoudenmire$ leveraged these properties of TNs. 
They applied the TT model to the handwritten digit classification task, demonstrating the impressive capabilities
\cite{stoudenmire2016schwab}.
In addition to the affinity of TNs for machine learning tasks, TNs have a unique architecture that enables efficient evaluation of multi-dimensional integrals.
The key to this efficiency is the commutative contractions of TNs, which allow the evaluation of multi-dimensional integrals with the sum and product operation of polynomially-sized tensors.
In contrast, MLPs present significant challenges for multi-dimensional integral calculations because their fully connected architecture and nonlinear activation functions prevent the application of commutative contractions during forward propagation.

For the evaluation of the integral between wavefunctions and operators written in TNs, the matrix product operator (MPO) \cite{pirvu2010verstraete} form has been introduced, which is a TT format operator representation and enables the evaluation of the integral in polynomial time provided that the wavefunction is represented by MPS or expanded by a polynomial number of configurations.
In this study, TT machine learning model convertible to MPO was applied to predict molecular PES. 
We also demonstrated its efficiency for many-body quantum simulations through phonon DMRG
calculations integrated with the trained TT model.

It is worth noting that while interatomic neural network potentials excel in scalability and generalization across diverse molecular systems, potentials based on SOP, HDMR and TN formalisms are primarily designed for specific molecular conformations comprising up to dozens of atoms. These approaches are particularly suited for elucidating spectroscopic properties, including absorption and emission spectra, proton tunneling effects, and ultrafast relaxation dynamics.

\section{Methods}
\subsection{Model Architecture}

In this paper, the dataset 
\begin{equation}
\mathcal{D} = 
\left\{
  \left(\mathbf{x}^{(k)}, V^{(k)}, \mathbf{F}^{(k)}\right)
  \middle| k=1, 2, \cdots, |\mathcal{D}|
\right\}
\end{equation}
contains $|\mathcal{D}|$ molecular mass-weighted coordinates $\mathbf{x}^{(k)} \in \mathbb{R}^{1\times n}$ (row-vector) and corresponding potential energies $V\left(\mathbf{x}^{(k)}\right) \in \mathbb{R}$ and analytical forces $\mathbf{F}\left(\mathbf{x}^{(k)}\right)=
-\frac{\partial V\left(\mathbf{x}^{(k)}\right)}{\partial \mathbf{x}} 
\in \mathbb{R}^{1\times n}$ where $n$ is the number of degrees of freedom of nuclear motion.
The batch index $(k)$ is omitted in the following description for simplicity.
We designate our model a neural network matrix product operator (NN-MPO), which is represented by the following equation:
\begin{eqnarray}
  \label{eq:nnmpo}
  {V}_{\text{NN-MPO}} (\mathbf{x})&=& \Phi(\mathbf{q})\mathbf{W} \\
  \mathbf{q} &=& \mathbf{x} U
\end{eqnarray}
where $U \in \text{St}(f,n) = \{U \in \mathbb{R}^{n\times f} | U^\top U = I_f\}$ is the orthogonal linear transformation matrix that maps the input mass-weighted coordinates $\mathbf{x}$ to the latent space coordinates $\mathbf{q} \in \mathbb{R}^{1\times f}$: 
\begin{equation}
  \label{eq:coordinator}
  \begin{bmatrix}
    q_1 & q_2 & \cdots & q_f 
  \end{bmatrix}
  = 
  \begin{bmatrix}
    x_1 & x_2 & \cdots & x_n
  \end{bmatrix}
  U.
\end{equation}
We refer to $U$ as $Coordinator$. 
Eq.~(\ref{eq:nnmpo}) can be interpreted as a sort of kernel method, where $\Phi$ serves as a design matrix and $\mathbf{W}$ as kernel weights.
$\Phi(\mathbf{q})$ is represented by tensor product basis:
\begin{equation}
  \Phi_{\rho_1\rho_2\cdots\rho_f}
  = \{\phi_{\rho_1}^{[1]}(q_1)\} \otimes \{\phi_{\rho_2}^{[2]}(q_2)\} 
  \otimes \cdots \otimes \{\phi_{\rho_f}^{[f]}(q_f)\}.
\end{equation}
Let the number of basis functions for every degree of freedom be $N$, $i.e.$ $\rho_i$ takes values $1, 2, \cdots, N$, resulting in a total number of expanded basis $N^f$.
The model can achieve high representational power by increasing $N$ to a sufficiently large value.
For simplicity, we rewrite the basis function $\phi_{\rho_i}^{[i]}(q_i)$ as $\phi_{\rho_i}$.
The choice of the basis function $\phi_{\rho_i}$ is crucial for the performance.
In the previous study for the handwritten digit classification \cite{stoudenmire2016schwab},
the authors employed
$\phi_{\rho_i}(x_i)=
\left[\cos\left(\frac{\pi}{2}x_i\right), 
      \sin\left(\frac{\pi}{2}x_i\right)\right]$
for the pixel value $x_i$.
However, this choice is not suitable for the molecular PES.
For instance, $Baranov$ and $Oseledets$ demonstrated the TT-based PES \cite{baranov2015oseledets} using $\phi_{\rho_i}$ as Chebyshev polynomial, 
trained through the combination of tensor cross interpolation \cite{oseledets2010tyrtyshnikov,oseledets2011oseledets} and Chebyshev approximation, which is an interpolation method for Chebyshev nodes.
This tensor train approach selects training data points dynamically through an interpolation process utilizing Chebyshev nodes, which are uniquely determined by specifying the input interval and polynomial order.
On the other hand, NN-MPO can be trained through gradient-based optimization, which does not limit the number and position of training data points.
Therefore, we can sample the training data points from the physically reasonable distribution.
Based on the strong prior knowledge of the PES structure, we chose the following basis function:
\begin{align}
  &\phi_{\rho_i}(q_i) = \notag \\
  &
  \label{eq:phi}
  \begin{cases}
    1 & \text{for $\rho_i=1$} \\
    1 - \exp\left(-\left(q_{\rho_i}^{(i)}\right)^2\right) 
    + \epsilon \left(q_{\rho_i}^{(i)}\right)^2 & 
    \text{for $2\leq \rho_i \leq \lceil \frac{N}{2}\rceil$} \\
    \frac{q_{\rho_i}^{(i)}}{1+\exp\left(-q_{\rho_i}^{(i)}\right)} 
    & \text{for $\lceil \frac{N}{2}\rceil < \rho_i \leq N$}
  \end{cases}
\end{align}
where 
\begin{equation}
  q_{\rho_i}^{(i)} = w_{\rho_i}^{(i)}\left(q_i-\bar{q}_{\rho_i}^{(i)}\right) 
  + b_{\rho_i}^{(i)}
\end{equation}
and $\bar{q}_{\rho_i}^{(i)} = \sum_j \bar{x}_{\rho_i}^{(j)} U_{ji}$. 
We randomly chose reference positions $\bar{x}_{\rho_i}^{(j)}$ from the training data.
It could be seemed that $\bar{q}_{\rho_i}^{(i)}$ is redundant, 
$-w_{\rho_i}^{(i)}\bar{q}_{\rho_i}^{(i)}$ is absorbed into $b_{\rho_i}^{(i)}$,
however, finding initial $b_{\rho_i}^{(i)}$ stable to the change of coordinator $ U $ is difficult.
Therefore, like a kernel method, we chose reference positions $\bar{x}_{\rho_i}^{(j)}$ from training data in advance and initialized $b_{\rho_i}^{(i)}=0$.
Besides $\mathbf{W}$, the traing parameters are $U$, $w_{\rho_i}^{(i)}$, and $b_{\rho_i}^{(i)}$.
In Eq.~(\ref{eq:phi}), the first activation function is nothing other than a constant function, 
which is inspired by the idea of HDMR: $n$-body function can be approximated by the sum of up to $k$-body functions, 
which $n-k$ variables of them are constant.
The second one, $1-\exp(-x^2) + \epsilon x^2$, was introduced by $Koch\;et\;al$ to the SOP neural network in the application to the excited state PES of \form molecule \cite{koch2019burghardt}, 
which is similar to a Gaussian function but moderately increases the quadratic term and may probably be suitable for the vibrational PES because it prohibits the nonphysical holes in the PES.
We set $\epsilon=0.05$, the same as the previous work.
The third one, $\frac{x}{1+\exp(-x)}$, is known as SiLU or $swish$ function \cite{ramachandran2017le}, a sophisticated activation function in a deep learning community, and we realized that this function is similar to Morse potential or Lennard-Jones potential, which may probably be suitable for the description of the dissociation limit.

\begin{figure}
  \centering
  \includegraphics[width=0.80\columnwidth]{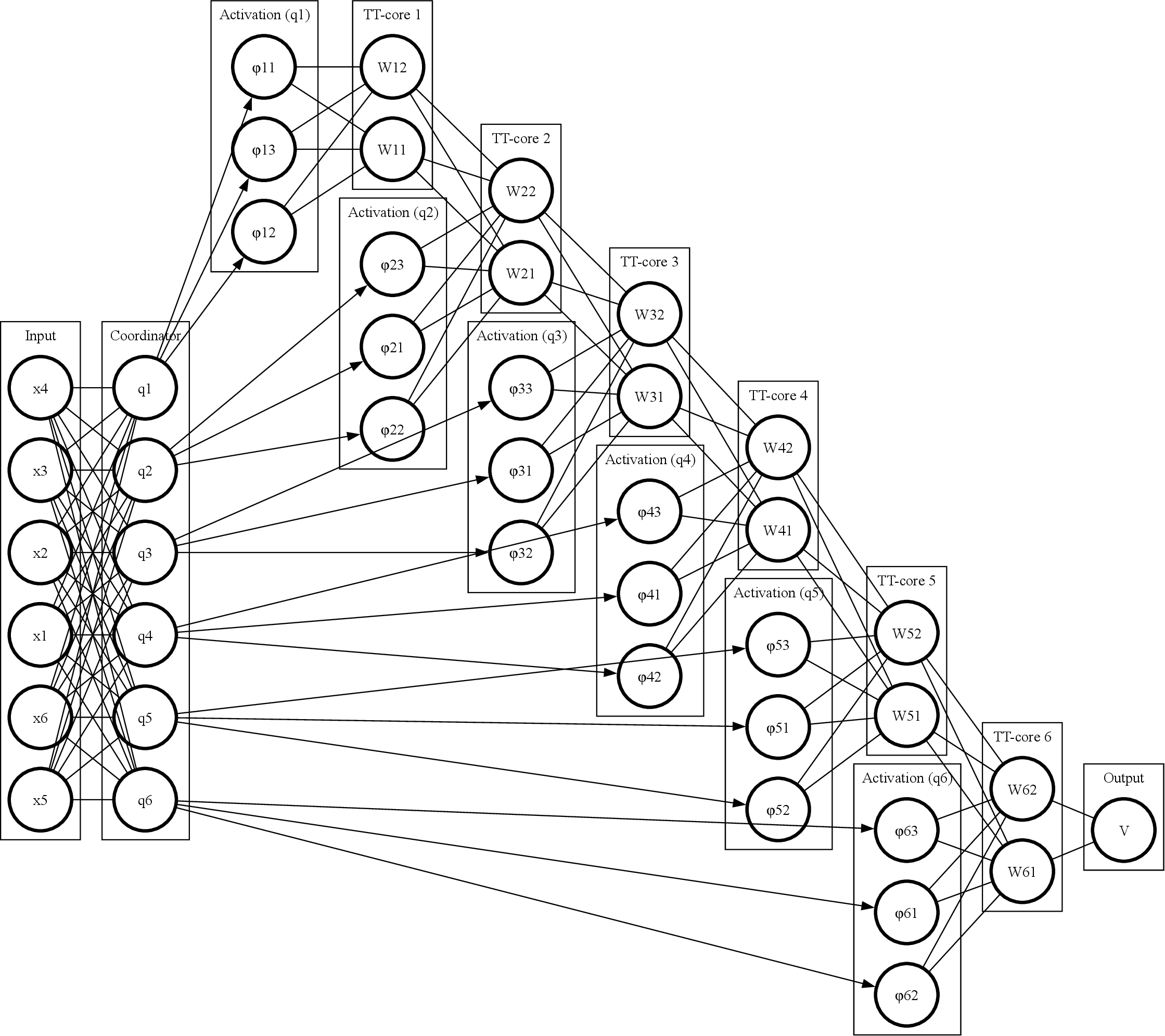}
  \caption{
  \label{fig:nnmpo}
  The conventional diagram in the deep learning community.
  This diagram indicates NN-MPO for $n=f=6$, $N=3$, and $M=2$.
}
\end{figure}

\begin{figure}
  \centering
  \includegraphics[width=0.80\columnwidth]{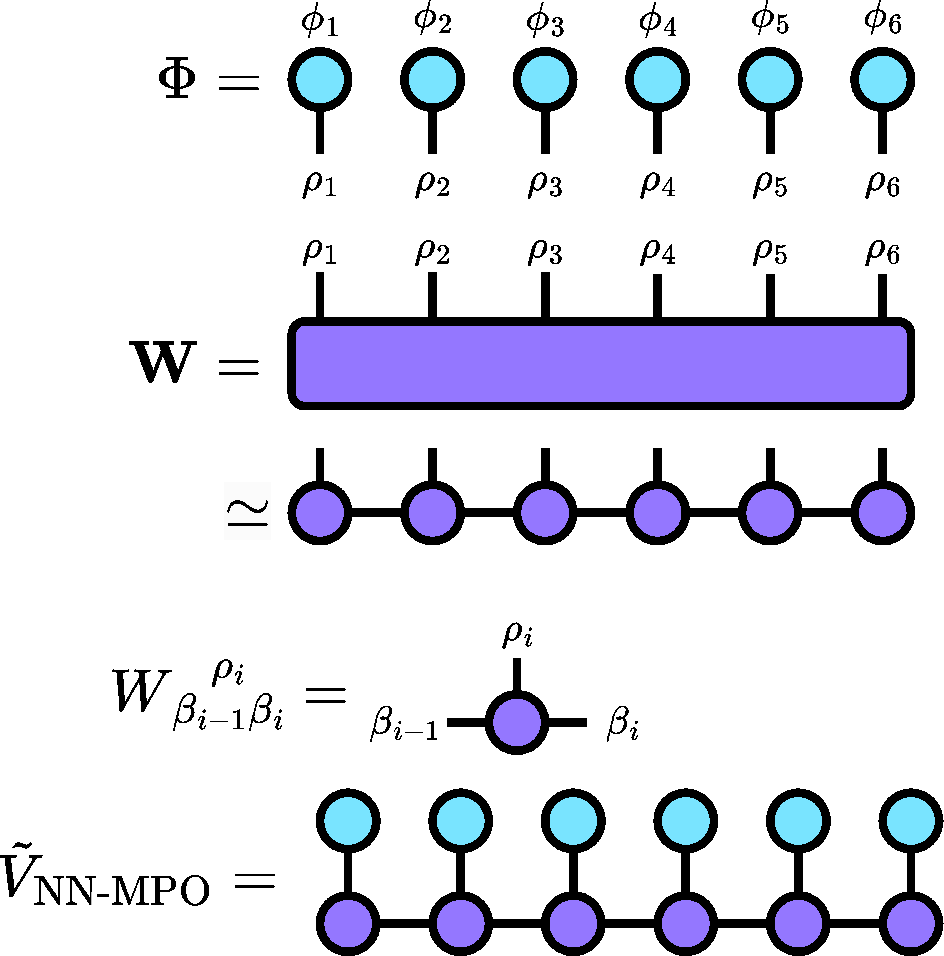}
  \caption{
  \label{fig:nnmpo-def}
  The conventional diagram in the tensor network community.
  This diagram indicates NN-MPO for $f=6$. Coordinator layer is omitted.
}
\end{figure}

Once the basis function $\Phi$ is defined, the weight $\mathbf{W}$ can be optimized.
While the size of $\mathbf{W}$ can naively reach $N^f$, we can reduce the number of parameters by using the TT structure:
\begin{equation}
  \label{eq:tn}
  \mathbf{W} = \mathbf{W}^{(1)}\mathbf{W}^{(2)}\cdots\mathbf{W}^{(f)}
\end{equation}
in other form,
\begin{equation}
  W_{\rho_1\rho_2\cdots\rho_f}
  = \sum_{\beta_{1}\beta_{2}\cdots\beta_{f-1}}
  W\substack{\rho_1\\1\beta_1}
  W\substack{\rho_2\\\beta_1\beta_2}
  \cdots
  W\substack{\rho_f\\\beta_{f-1}1}
\end{equation}
where each ``core'' tensor $W\substack{\rho_i\\\beta_{i-1}\beta_i} \in \mathbb{R}^{M_{i-1}\times N\times M_i}$ can achieve the low-rank approximation, which enables not only the reduction of parameter but also the mitigation of complexity errors such as overfitting.
The connecting index $\beta$ takes $\beta=1, 2, \cdots, M$ where $M$ is called bond dimension, link dimension, or TT-rank.
The following equations write the full formulation of NN-MPO:
\begin{align}
  &V_{\text{NN-MPO}}(\mathbf{x}) = \widetilde{V}_{\text{NN-MPO}}(\mathbf{q}) \notag \\
  &=
  \label{eq:nnmpo-full}
  \sum_{\substack{\rho_1,\rho_2,\cdots\rho_f\\
        \beta_1,\beta_2,\cdots\beta_{f-1}}}
  \phi_{\rho_1}(q_1) \cdots \phi_{\rho_f}(q_f)
  W\substack{\rho_1\\1\beta_1}W\substack{\rho_2\\\beta_1\beta_2}
  \cdots W\substack{\rho_f\\\beta_{f-1}1}.
\end{align}
The corresponding diagrams of the architecture are shown in Fig.~\ref{fig:nnmpo} and Fig.~\ref{fig:nnmpo-def}.
\subsection{Optimization}
\def\BlockL{\varPhi^{[:i-1]}_{p\beta_{i-1}}}
\def\BlockR{\varPhi^{[i+2:]}_{p\beta_{i+1}}}
\def\BlockAll{\varPhi\substack{p\\\rho_i\rho_{i+1}\\\beta_{i-1}\beta_{i+1}}}
\def\TwoDot{B\substack{\rho_i\rho_{i+1}\\\beta_{i-1}\beta_{i+1}}}
\def\OneDot{C\substack{\rho_i\\\beta_{i-1}\beta_{i}}}

We defined the loss function $\mathcal{L}$ as a sum of energy and force mean squared errors (MSE).
\begin{equation}
  \label{eq:loss}
  \mathcal{L} = 
  \mathcal{L}_{\text{energy}} +
  \mathcal{L}_{\text{force}} 
\end{equation}
\begin{equation}
  \label{eq:loss-energy}
  \mathcal{L}_{\text{energy}} = 
  \frac{1}{|\mathcal{D}|}\sum_{\mathbf{x}, V\in\mathcal{D}}
  \frac{1}{2}
  \left\| 
    V_{\text{NN-MPO}}(\mathbf{x}) - V
  \right\|^2
\end{equation}
\begin{align}
  \label{eq:loss-force}
  \mathcal{L}_{\text{force}} 
  &= 
  \frac{1}{|\mathcal{D}|}\sum_{\mathbf{x}, \mathbf{F}\in\mathcal{D}}
  \frac{1}{2}
  \left\| 
    - \frac{\partial V_{\text{NN-MPO}}(\mathbf{x})}{\partial \mathbf{x}}
    - \mathbf{F}
  \right\|^2
  \\
  \label{eq:loss-force-latent}
  &=
  \frac{1}{|\mathcal{D}|}\sum_{\mathbf{x}, \mathbf{F}\in\mathcal{D}}
  \frac{1}{2}
  \left\| 
    - \frac{\partial \tilde{V}_{\text{NN-MPO}}(\mathbf{q})}{\partial \mathbf{q}}
    - \mathbf{F}U
  \right\|^2
\end{align}
The gradient of $w_{\rho_i}^{(i)}, b_{\rho_i}^{(i)}$ and $U$ to the loss function $\mathcal{L}$ can be evaluated by the automatic differentiation facilitated by the deep learning framework. In our implementation, \texttt{JAX} library was used \cite{jax2018github}.
These parameters were updated by the Adam optimization \cite{kingma2017ba}. 
In addition, we used QR decomposition to retract $U$ onto the Stiefel manifold to keep the orthogonality for each step \cite{bonnabel2013bonnabel,becigneul2019ganea}.
We referred to the implementation of the Riemannian Adam optimization algorithm in \texttt{geoopt} library \cite{geoopt2020kochurov}.
The computational complexity of the Coordinator layer optimization scales at least as $\mathcal{O}(f^3)$ due to QR decomposition and gradient evaluation. However, this cost is typically less significant than the sweeping optimization described below for two key reasons: (1) basis function optimization can be performed with loose convergence criteria, and (2) the matrix $U$ can be implemented in a sparse block-diagonal form, as coordinate rotation can be limited to physically coupled coordinate pairs when extending to larger systems.

\begin{figure}
  \includegraphics[width=0.50\columnwidth]{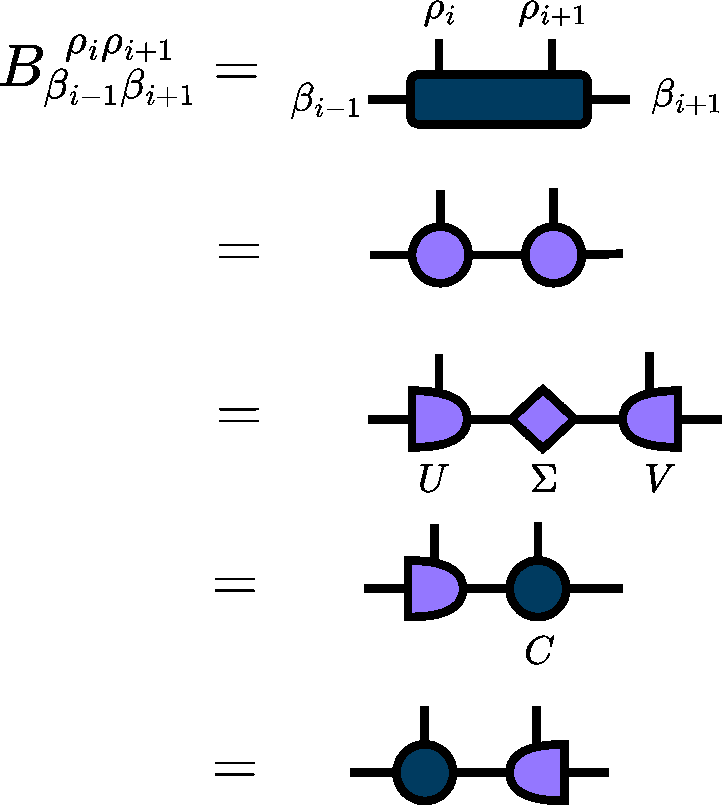}
  \caption{
  \label{fig:two-dot-svd}
  The diagram of two-dots and one-dot center core in TT, both of which are filled with navy color.
  }
\end{figure}

The optimization scheme of TT core tensor $W\substack{\rho_i\\\beta_{i-1}\beta_i}$ is
different from the conventional neural network optimization because TN has gauge freedom, which means that TN can return the same value from the contraction of different tensors, for instance, 
$A_{ij}=\sum_k U_{ik}\Sigma_{kk}V_{kj} = \sum_k U_{ik} C_{kj} = \sum_k D_{ik}V_{kj}$. 
The DMRG algorithm successfully imposed the canonical gauge form on the MPS and updated each core tensor from left to right and right to left (a strategy called $sweeping$). We employed the same approach for the TT core optimization. 
It is helpful to introduce the following gauge fixed notation:
\begin{align}
  \label{eq:two-dot-gauge}
  \TwoDot
  &:= \sum_{\beta_i} 
  W\substack{\rho_i\\\beta_{i-1}\beta_i}
  W\substack{\rho_{i+1}\\\beta_i\beta_{i+1}}
  \\
  \label{eq:two-dot-svd}
  &= \sum_{\beta_i} 
  U\substack{\rho_i\\\beta_{i-1}\beta_i}
  \Sigma\substack{\beta_i\;\\\;\beta_i}
  V\substack{\rho_{i+1}\\\beta_i\beta_{i+1}}
  \\
  &= \sum_{\beta_i} 
  U\substack{\rho_i\\\beta_{i-1}\beta_i}
  C\substack{\rho_{i+1}\\\beta_i\beta_{i+1}}
  \\
  &= \sum_{\beta_i} 
  C\substack{\rho_i\\\beta_{i-1}\beta_i}
  V\substack{\rho_{i+1}\\\beta_i\beta_{i+1}}.
\end{align}
Eq~(\ref{eq:two-dot-svd}) is the singular value decomposition (SVD) of the matrix with row $\rho_i\otimes\beta_{i-1}$ and column $\rho_{i+1}\otimes\beta_{i+1}$.
The diagram corresponding to these equations is shown in Fig.~\ref{fig:two-dot-svd}.
By repeating the SVD of the TT core tensor,
The following gauge-fixed equation rewrites TT:
\begin{equation}
\begin{aligned}
  \label{eq:tt-svd}
  &
  \sum_{\beta_1\cdots\beta_{f-1}}
  W\substack{\rho_1\\1\beta_1}
  W\substack{\rho_2\\\beta_1\beta_2}
  \cdots
  W\substack{\rho_f\\\beta_{f-1}1}
  \\
  &= 
  \sum_{\beta_1\cdots\beta_{f-1}}
  C\substack{\rho_1\\1\beta_1}
  V\substack{\rho_2\\\beta_1\beta_2}
  \cdots
  V\substack{\rho_{f}\\\beta_{f-1}1}
  \\
  &=
  \sum_{\beta_2\cdots\beta_{f-1}}
  B\substack{\rho_1\rho_2\\1\beta_2}
  V\substack{\rho_3\\\beta_2\beta_3}
  \cdots
  V\substack{\rho_{f}\\\beta_{f-1}1}
  \\
  &=
  \sum_{\beta_1\cdots\beta_{f-1}}
  U\substack{\rho_1\\1\beta_1}
  C\substack{\rho_2\\\beta_1\beta_2}
  V\substack{\rho_3\\\beta_2\beta_3}
  \cdots
  V\substack{\rho_{f}\\\beta_{f-1}1}
  \\
  &\vdots
\end{aligned}
\end{equation}
Then, we can minimize the loss function $\mathcal{L}$ to 
$\TwoDot$ (as well as $\OneDot$).
The force loss function $\mathcal{L}_{\text{force}}$ can be further simplified.
The predicted force in Eq~(\ref{eq:loss-force-latent}) is $-\frac{\partial \tilde{V}_{\text{NN-MPO}}(\mathbf{q})}{\partial \mathbf{q} }
= \mathbf{W} \left(-\frac{\partial \Phi(\mathbf{q})}{\partial \mathbf{q}}\right)$
where
\begin{equation}
  \begin{aligned}
  &- \frac{\partial \Phi(\mathbf{q})}{\partial \mathbf{q}}
  \\
  &= 
  \begin{bmatrix}
    - \frac{\partial \Phi(\mathbf{q})}{\partial q_1} 
    & - \frac{\partial \Phi(\mathbf{q})}{\partial q_2}
    & \cdots
    & - \frac{\partial \Phi(\mathbf{q})}{\partial q_f}
  \end{bmatrix}
  \\
  &=
  \begin{bmatrix}
    \left\{- \frac{\partial \phi_{\rho_1}}{\partial q_1}\right\} & \otimes & \left\{\phi_{\rho_2}                                \right\} & \otimes & \cdots & \otimes & \left\{\phi_{\rho_f} \right\}\\
    \left\{\phi_{\rho_1}                                \right\} & \otimes & \left\{- \frac{\partial \phi_{\rho_2}}{\partial q_2}\right\} & \otimes & \cdots & \otimes & \left\{\phi_{\rho_f} \right\}\\
                                                                 &         &                                                              & \vdots  &        &         &                      \\
    \left\{\phi_{\rho_1}                                \right\} & \otimes & \left\{\phi_{\rho_2}                                \right\} & \otimes & \cdots & \otimes & \left\{-\frac{\partial \phi_{\rho_f}}{\partial q_f}\right\} 
  \end{bmatrix}^\top.
  \end{aligned}
\end{equation}
To simplify the loss function, we redefine the database $\mathcal{D}^\prime$ which contains $|\mathcal{D}^\prime|=|\mathcal{D}|\times (f+1)$ data.
The index of the data point is redefined as
\begin{equation}
  \{p\} := \{k\} \otimes \{r\}
\end{equation}
where $k=1,2,\cdots,|\mathcal{D}|$ is the index of the original data point and $r=1,2,\cdots,f+1$ is the index of the concatenated array of $\mathbf{F}$ and $V$.
The input and output set is redefined as
\begin{equation}
  \varphi_{\rho_i}^{p} :=
  \bar{\varphi}_{r\rho_i}^{(k)} :=
  \begin{cases}
    -\frac{\partial \phi_{\rho_i}\left(q_i^{(k)}\right)}{\partial q_i} & \text{for $r=i$} \\
    \phi_{\rho_i}\left(q_i^{(k)}\right) & \text{otherwise} 
  \end{cases}
\end{equation}
\begin{equation}
  y_p :=
  \bar{y}_r^{(k)} :=
  \begin{cases}
    \sum_i \mathbf{F}^{(k)}_i U_{ir} & \text{for $1 \leq r \leq f$} \\
    V^{(k)} & \text{for $r=f+1$}
  \end{cases}
\end{equation}
\begin{figure}
  \includegraphics[width=0.80\columnwidth]{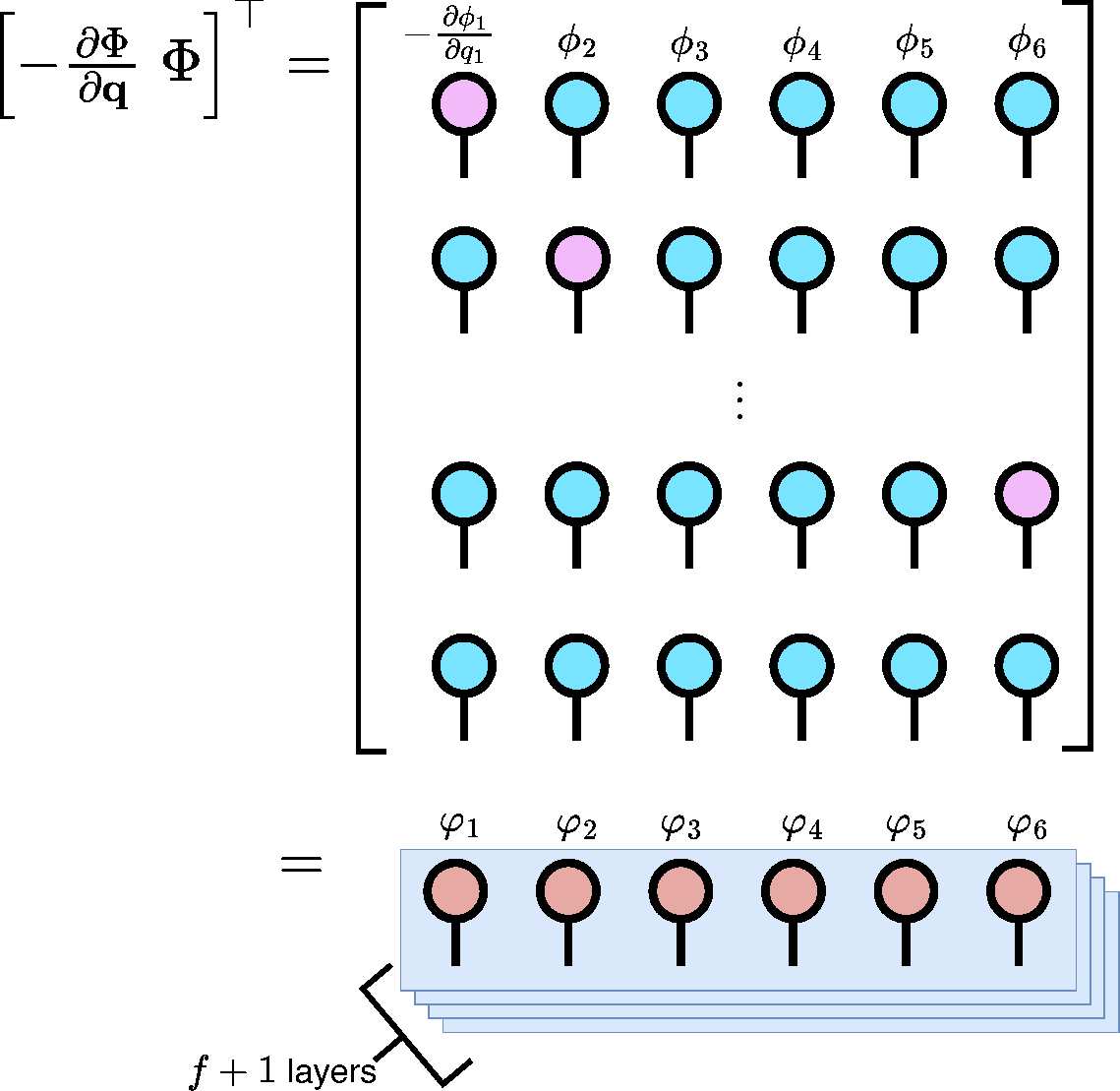}
  \caption{
  \label{fig:force}
  The diagram of basis concatenated force and energy.
  }
\end{figure}
Furthermore, we introduce the following notation:
\begin{equation}
  \BlockL :=
  \sum_{\substack{\rho_1\cdots\rho_{i-1}\\
      \beta_1\cdots\beta_{i-2}}}
  U\substack{\rho_1\\1\beta_1}
  \cdots
  U\substack{\rho_{i-1}\\\beta_{i-2}\beta_{i-1}}
  \varphi_{\rho_1}^{p}
  \cdots
  \varphi_{\rho_{i-1}}^{p}
\end{equation}
\begin{equation}
  \BlockR := 
  \sum_{\substack{\rho_{i+2}\cdots\rho_f\\
      \beta_{i+2}\cdots\beta_{f-1}}} 
  V\substack{\rho_{i+2}\\\beta_{i+1}\beta_{i+2}}
  \cdots
  V\substack{\rho_f\\\beta_{f-1}1}
  \varphi_{\rho_{i+2}}^{p}
  \cdots
  \varphi_{\rho_f}^{p}
\end{equation}
\begin{equation}
  \BlockAll
  :=
  \BlockL
  \varphi_{\rho_i}^{p}
  \varphi_{\rho_{i+1}}^{p}
  \BlockR
\end{equation}
Then, the total loss function becomes
\begin{equation}
  \label{eq:loss-tt}
    \mathcal{L} = 
    \frac{1}{|\mathcal{D}|}
    \sum_p
    \bar{\mathcal{L}}
    \left(
      \TwoDot; \BlockAll, y_p
    \right)
\end{equation}
where
\begin{equation}
\begin{aligned}
    & \bar{\mathcal{L}}
    \left(
      \TwoDot; \BlockAll, y_p
    \right)
    \\
    &= 
    \frac{1}{2}
    \left\| 
    \sum_{\substack{\rho_i\rho_{i+1}\\\beta_{i-1}\beta_{i+1}}}
    \BlockAll
    \TwoDot
    - y_p
    \right\|^2.
\end{aligned}
\end{equation}
We can minimize Eq.~(\ref{eq:loss-tt}) in either two ways.
One way is to use the analytical gradient of 
$\TwoDot$:
\begin{equation}
  \frac{\partial \bar{\mathcal{L}}}{\partial B}
  = 
  \left(
  \sum_{\substack{\rho_i\rho_{i+1}\\\beta_{i-1}\beta_{i+1}}}
  \BlockAll\TwoDot- y_p
  \right)
  \BlockAll
\end{equation}
and update $\TwoDot$ by the gradient-based optimization algorithm such as Adam.
This approach is suitable when the batch size $|\mathcal{D}^\prime|$ and the number of parameters $M^2N^2$ are significant or gradual changes of TT required.
The other way is regarding the minimization problem as a quadratic form problem which minimizes the following function:
\begin{equation}
  f\left(\boldsymbol{x}\right) = \frac{1}{2}\boldsymbol{x}^\top A \boldsymbol{x} - \boldsymbol{x}^\top b
  \label{eq:quadratic-form}
\end{equation}
where $\boldsymbol{x}$ is the vectorized $B\substack{\rho_i\rho_{i+1}\\\beta_{i-1}\beta_{i+1}}$,
Hessian matrix 
$A=\sum_p\left(
  \varPhi\substack{p\\\rho_i^\prime\rho_{i+1}^\prime\\\beta_{i-1}^\prime\beta_{i+1}^\prime}
\right)^\top \BlockAll$ 
and $b=\sum_p\left(\BlockAll\right)^\top y_p$.
This problem is solved by the conjugate gradient (CG) method, which is detailed in the appendix along with its computational complexity.
This approach is suitable when the batch size $|\mathcal{D}^\prime|$ and the number of parameters $M^2N^2$ are small and rapid convergence to the fixed basis function is required.

Once the TT core tensor $B\substack{\rho_i\rho_{i+1}\\\beta_{i-1}\beta_{i+1}}$ is optimized, SVD is executed to truncate the bond dimension to $M$.
Then, the center core tensor is swept to the neighboring core by gauge transformation in Eq.~(\ref{eq:two-dot-gauge}), and repeat these procedures until the optimal value is obtained.
In addition, by replacing $\TwoDot$ with $\OneDot$ and $\BlockR$ with $\varPhi^{[i+1:]}_{p\beta_{i}} = \sum_{\substack{\rho_{i+1}\\\beta_{i+1}}} 
\varphi^{p}_{\rho_{i+1}}V\substack{\rho_{i+1}\\\beta_i\beta_{i+1}}\BlockR$,
one-dot core tensor $C\substack{\rho_i\\\beta_{i-1}\beta_{i}}$ 
can be optimized in the same way except for the SVD truncation.
We observed that until the basis function $\mathbf{\Phi}$ converged to the optimal value, the TT should be gradually optimized, but once basis function $\mathbf{\Phi}$ converged, the TT optimization should be done by the CG method with tight convergence criteria. 
In addition, once the bond dimension $M$ reaches the maximum value defined in advance, from the next sweeping, the optimization should be executed in the one-dot form $C\substack{\rho_i\\\beta_{i-1}\beta_{i}}$, which is free from the truncation error.
In general, increasing the bond dimension $M$ enhances the representational power, but it also increases the risk of overfitting. 
Therefore, the bond dimension $M$ is a crucial hyperparameter in TT optimization.
\begin{figure}
  \includegraphics[width=0.80\columnwidth]{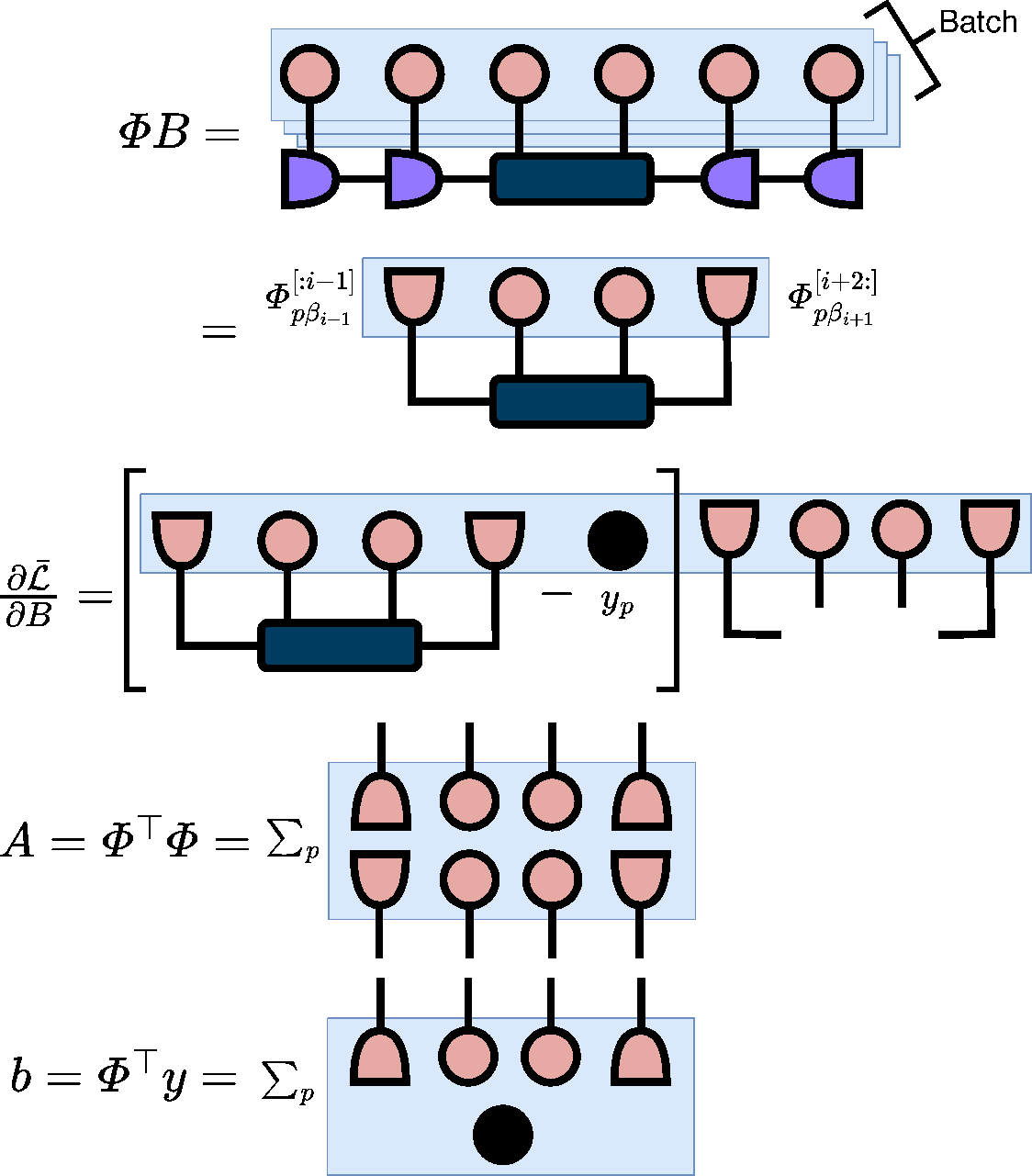}
  \caption{
  \label{fig:block-nnmpo}
  The diagram of tensor network blocks used for the optimization.
  }
\end{figure}
\subsection{Multi-dimensional integral}
NN-MPO demonstrates sufficient capability as a high-precision energy predictor.
However, its true strength lies in its compatibility with multi-dimensional integrals, such as the evaluation of expectation values.
This section shows how the multi-dimensional integral between the wavefunction basis and the NN-MPO can be evaluated in polynomial time.
We consider the wavefunction in $f$-dimensional space $\mathbf{q} \in \mathbb{R}^{1\times f}$ written as
\begin{equation}
  \Ket{\Psi} = 
  \sum_{\boldsymbol{\sigma}} 
  A_{\boldsymbol{\sigma}}
  \Ket{\sigma_1\left(q_1\right)}
  \Ket{\sigma_2\left(q_2\right)}
  \cdots\Ket{\sigma_f\left(q_f\right)}
\end{equation} 
where $A_{\boldsymbol{\sigma}}$ is the coefficient of each configuration $\boldsymbol{\sigma}$.
$A_{\boldsymbol{\sigma}}$ can be a linear combination of state vectors, mean-field wavefunction, or MPS.

We consider the Hamiltonian represented by
\begin{align}
  H &= 
  \label{eq:hamiltonian-mw}
  \sum_{i=1}^{n} -\frac{\hbar^2}{2}\frac{\partial^2}{\partial x_i^2} + V(x_1, x_2, \cdots, x_n)
  \\
  &\simeq
  \label{eq:hamiltonian-mw-hidden}
  \sum_{i=1}^{f} -\frac{\hbar^2}{2}\frac{\partial^2}{\partial q_i^2} + V(q_1, q_2, \cdots, q_f)
  \\
  &\simeq
  \label{eq:hamiltonian-mw-hidden-nnmpo}
  \sum_{i=1}^{f} -\frac{\hbar^2}{2}\frac{\partial^2}{\partial q_i^2} +
  \widetilde{V}_{\text{NN-MPO}}(q_1, q_2, \cdots, q_f).
\end{align}
The first line, Eq~(\ref{eq:hamiltonian-mw}), is the Hamiltonian in the mass-weighted coordinate space, in which atomic masses are absorbed into the positions $x_i$.
The second line, Eq~(\ref{eq:hamiltonian-mw-hidden}), is the Hamiltonian in the latent coordinate space $\mathbf{q}$, which is exact when $n=f$.
Since the coordinate transformation is linear and orthogonal, the kinetic terms are kept in a simple form.
When $n>f$, the equation represents an approximation where degrees of freedom are reduced through energy and one-dimensional entanglement-aware dimensionality reduction, provided the Coordinator is properly optimized.
This dimensionality reduction could be a sufficient choice for accurate prediction of quantum mechanical energetics,
including vibrational eigenenergies and energy redistribution phenomena.
The third line, Eq~(\ref{eq:hamiltonian-mw-hidden-nnmpo}), is the Hamiltonian approximated by NN-MPO.

\begin{figure}
  \centering
  \includegraphics[width=0.80\columnwidth]{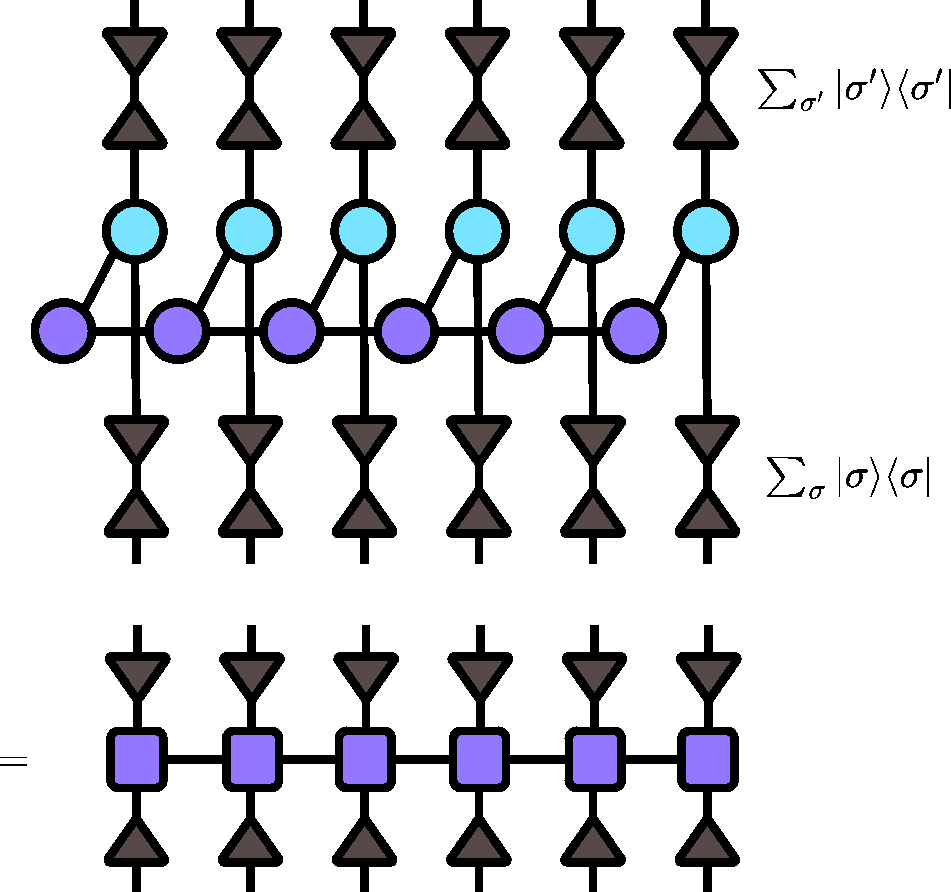}
  \caption{
  \label{fig:nnmpo-mpo}
  Conversion process from NN-MPO to MPO.
}
\end{figure}

While the integral between kinetic terms and wavefunction is easily evaluated because it has a SOP form, the integral between the potential term and wavefunction is not trivial.
To evaluate the integral, we need to prepare the potential operator $\hat{V}$, derived from inserting the projection operator 
\begin{equation}
  \hat{V} = \sum_{\boldsymbol{\sigma}^\prime, \boldsymbol{\sigma}} 
  \Ket{\boldsymbol{\sigma}^\prime}
  \Braket{\boldsymbol{\sigma}^\prime|\tilde{V}_{\text{NN-MPO}}|\boldsymbol{\sigma}}
  \Bra{\boldsymbol{\sigma}}
\end{equation}
where 
\begin{equation}
  \Ket{\boldsymbol{\sigma}} = 
  \Ket{\sigma_1\left(q_1\right)}
  \Ket{\sigma_2\left(q_2\right)}
  \cdots\Ket{\sigma_f\left(q_f\right)}.
\end{equation}
At first, we evaluate one-dimensional integrals between potential basis $\phi_{\rho_i}(q_i)$ and wavefunction basis $\sigma_{i}(q_i)$:
\begin{equation}
  I\substack{\sigma_i^\prime \\ \rho_i \\ \sigma_i} = 
  \Braket{\sigma_i^\prime(q_i)|\phi_{\rho_i}(q_i)|\sigma_i(q_i)}.
\end{equation}
The total cost of evaluating all $I\substack{\sigma_i^\prime \\ \rho_i \\ \sigma_i}$ is $\mathcal{O}\left(d^2NfK\right)$, where $d$ is the number of wavefunction basis, $N$ is the number of potential basis, $f$ is the degrees of freedom, and $K$ is the cost of one-dimensional integral.
Next, we contract TT-core tensor $W\substack{\rho_i\\\beta_{i-1}\beta_i}$ and $I\substack{\sigma_i^\prime \\ \rho_i \\ \sigma_i}$:
\begin{equation}
  \mathcal{W}\substack{\sigma_i^\prime\\\beta_{i-1}\beta_i\\\sigma_i} =
  \sum_{\rho_i} 
  W\substack{\rho_i\\\beta_{i-1}\beta_i} 
  I\substack{\sigma_i^\prime \\ \rho_i \\ \sigma_i}.
\end{equation}
This contraction costs $\mathcal{O}\left(M^2d^2Nf\right)$ 
where $M$ is the bond dimension of TT.
Finally, we have the MPO of the potential operator:
\begin{equation}
  \hat{V} = 
  \sum_{\boldsymbol{\sigma}^\prime, \boldsymbol{\sigma}} 
  \Ket{\boldsymbol{\sigma}^\prime}
  \sum_{\beta_1\cdots\beta_{f-1}}
  \mathcal{W}\substack{\sigma_1^\prime\\1\beta_1\\\sigma_1}
  \mathcal{W}\substack{\sigma_2^\prime\\\beta_1\beta_2\\\sigma_2}
  \cdots
  \mathcal{W}\substack{\sigma_f^\prime\\\beta_{f-1}1\\\sigma_f}
  \Bra{\boldsymbol{\sigma}}.
\end{equation}
The conversion process is shown in Fig.~\ref{fig:nnmpo-mpo}.

By contracting from the left or right terminal site, the total evaluation cost of $\Braket{\boldsymbol{\sigma}^\prime|\hat{V}|\boldsymbol{\sigma}}$ reaches $\mathcal{O}\left(M^2f\right)$ when wavefunction basis is orthogonal.
The evaluation costs of mean-field operator $\Braket{\boldsymbol{\sigma}^{(i)\prime}|\hat{V}|\boldsymbol{\sigma}^{(i)}}$, where $\Ket{\boldsymbol{\sigma}^{(i)}} = \Ket{\sigma_1\cdots\sigma_{i-1}\sigma_{i+1}\cdots\sigma_f}$,
is also $\mathcal{O}\left(M^2f\right)$.
As long as the wavefunction is represented by MPS or expanded by a polynomial number of configurations, the multi-dimensional integral can be evaluated in polynomial time.

\section{Results and Discussion}
\subsection{Dataset sampling}
\begin{figure}
  \centering
  \includegraphics[width=0.9\columnwidth]{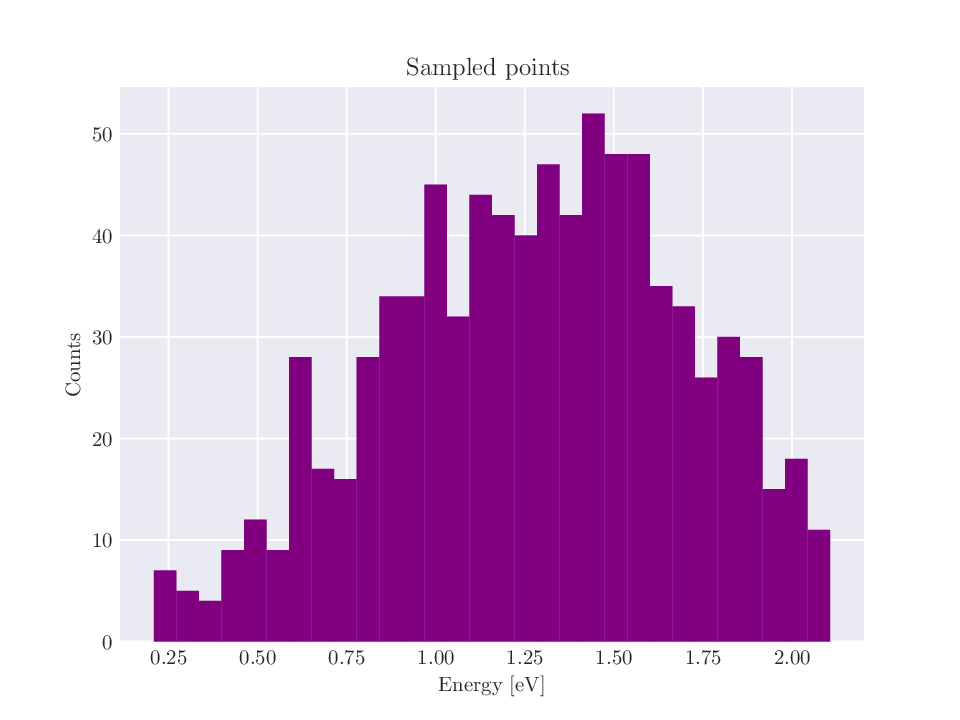}
  \caption{
  \label{fig:test_hist}
  Energy distribution of 839 test data points for \form molecule.
  Energies are shifted by the energy at the equilibrium structure.
  }
\end{figure}

To demonstrate the capability of the prediction and the compatibility with multi-dimensional integrals of NN-MPO, we show the application to 6-dimensional space spanned by all normal modes of formaldehyde \form as a test case.
\form is a typical benchmark molecule for both full-quantum approach and machine learning potential \cite{koch2019burghardt, kamath2018manzhos,manzhos2021carrington,koner2020meuwly} due to its strong anharmonicity and many-body correlation.
In this work, the input to the model is a position represented by six normal coordinates, and the output is a ground state energy at that position.
Energies and forces were computed by density functional theory (DFT) with the B3LYP functional and the SVP basis set with JK-fit approximation implemented in the BAGEL program package \cite{shiozaki2018shiozaki}.
Samplings were performed from inverted potential probability density function \cite{ku2019manzhos}
\begin{equation}
  \label{eq:potential-pdf}
  P(V) \propto \max\left(
    \frac{V_{\text{max}} + \Delta - V}{V_{\text{max}}  + \Delta}
    ,0
  \right)
\end{equation}
where we set $V_{\text{max}}=17,000$ \kayser and $\Delta=500$ \kayser, which are the same values as previous work \cite{kamath2018manzhos}.
From this distribution, in this work, 625 training data points and DFT energies up to $V_{\text{max}}$ were sampled by Metropolis-Hastings algorithm while previous work used analytical SOP model potential \cite{handy1997demaison} spanned by six internal coordinates and 
sampled by $quasi$ Monte Carlo method in the hypercuboid \cite{kamath2018manzhos}.
We should note that any function in SOP form can be exactly encoded into NN-MPO by choosing the appropriate basis in the same way as general MPO does \cite{ren2020shuai,crosswhite2008bacon}.
The sampled test data are shown in Fig.~\ref{fig:test_hist}.
\subsection{Training results}
\begin{figure}
  \centering
  \includegraphics[width=0.9\columnwidth]{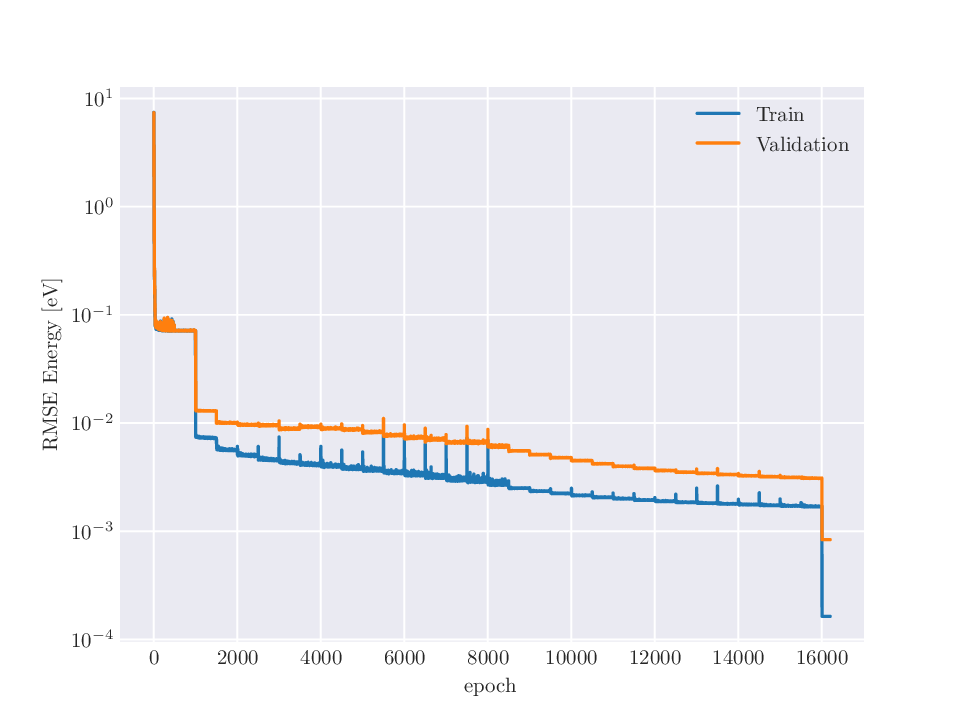}
  \caption{
  \label{fig:nnmpo-trace}
    Training trace of the NN-MPO ($M=14$, $N=21$) for \form molecule.
    For the first 1,000 epochs, basis and TT were optimized iteratively in short epochs.
    During 1,000-16,000 epochs, sweeping was executed every 500 epochs.
    Final drastic convergence was achieved by sweeping using the CG method.
  }
\end{figure}

\begin{figure}
  \centering
  \includegraphics[width=0.9\columnwidth]{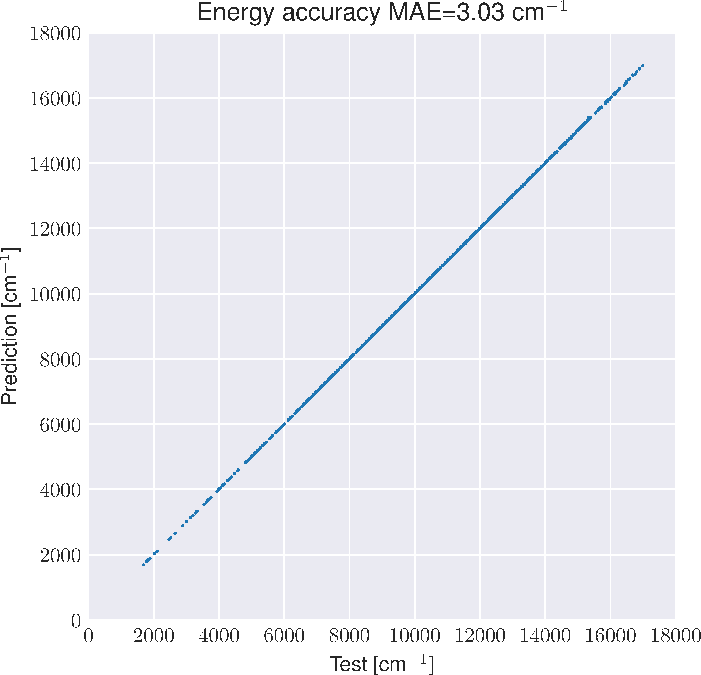}
  \caption{
  \label{fig:nnmpo-scatter}
    Scatter plot of the predicted energy by NN-MPO with $N=21$ and $M=14$ and test energy for \form molecule.
  }
\end{figure}

\begin{figure}
  \centering
  \includegraphics[width=0.9\columnwidth]{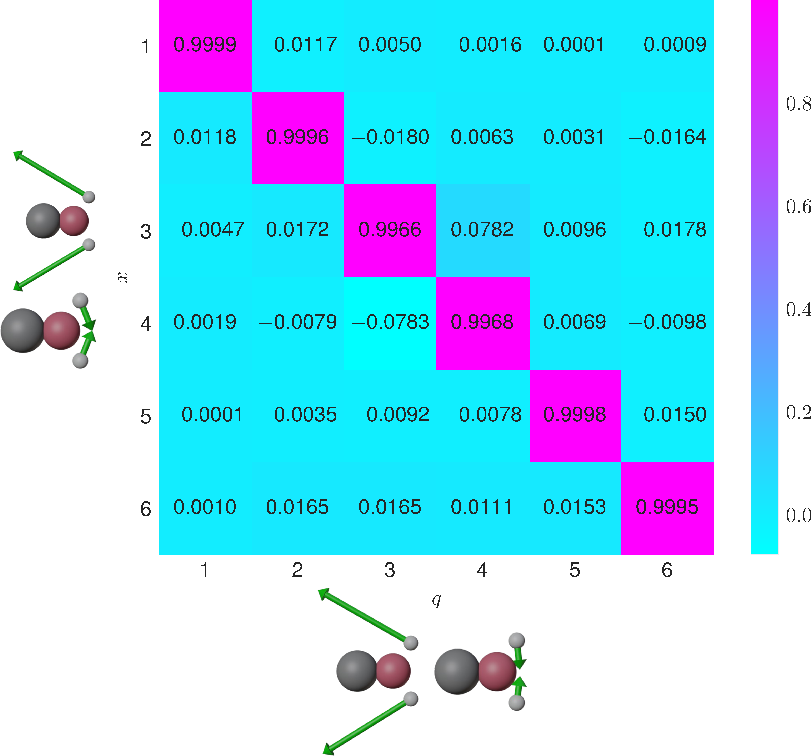}
  \caption{
  \label{fig:nnmpo-coordinator}
    Coordinator $U$ of NN-MPO for the normal modes of \form molecule.
    Row and column indices are the initial normal modes $x$ and the latent modes $q$, respectively.
  }
\end{figure}

For our model, we set the input dimension $n=6$, latent dimension $f=6$, initial bond dimension $M=2$, and systematically varied both the basis function count $N$ and maximum bond dimension $M$ to confirm the performance.
The order of the input normal coordinates was assigned in the frequency order given by 
harmonic oscillator (HO) approximation, which represents the PES as a quadratic form obtained by diagonalizing the Hessian matrix at the equilibrium geometry.
\begin{table}
  \centering
  \caption{\label{tab:rmse}
  RMSE of energy in meV for the validation set. (w) and (w/o) denote cases with and without the Coordinator layer, respectively.
  Values may vary depending on the random number generator seed.
  The left table shows just $NM^2$ for each case.
  }
  \begin{tabular}{|c|c|c|ccc|}
    \hline
    \multicolumn{3}{|c|}{} & \multicolumn{3}{c|}{Number of basis $N$} \\
    \cline{3-6}
    \multicolumn{2}{|c|}{} & $U$ & 11 & 21 & 41 \\
    \hline
    \multirow{8}{*}{\rotatebox{90}{Bond dimension $M$}} 
    & \multirow{2}{*}{4} & (w) & 29.736 & 31.043 & 28.207 \\
    & & (w/o) & 40.319 & 41.251 & 39.043 \\
    \cline{2-6}
    & \multirow{2}{*}{7} & (w) & 7.973 & 9.003 & 9.841 \\
    & & (w/o) & 9.177 & 9.681 & 13.667 \\
    \cline{2-6}
    & \multirow{2}{*}{14} & (w) & 1.426 & 0.836 & 1.028 \\
    & & (w/o) & 1.379 & 1.082 & 0.919 \\
    \cline{2-6}
    & \multirow{2}{*}{28} & (w) & 2.109 & 3.151 & 2.213 \\
    & & (w/o) & 2.911 & 1.755 & 1.743 \\
    \hline
  \end{tabular}
  \quad
  \begin{tabular}{|c|c|ccc|}
    \hline
    \multicolumn{2}{|c|}{} & \multicolumn{3}{c|}{$N$} \\
    \cline{3-5}
    \multicolumn{2}{|c|}{$NM^2$} & 11 & 21 & 41 \\
    \hline
    \multirow{4}{*}{\vspace{0.5em}$M$\vspace{0.5em}} 
    & 4 & 176 & 336 & 656 \\
    & 7 & 539 & 1,029 & 2,009 \\
    & 14 & 2,156 & 4,116 & 8,036 \\
    & 28 & 8,624 & 16,464 & 32,144 \\
    \hline
  \end{tabular}
\end{table}
The root mean square error (RMSE) of energy for validation set is shown in Table~\ref{tab:rmse}, 
which demonstrates that increasing $N$ and $M$ does not consistently improve results, as enhanced representational power can lead to overfitting. 
The relationship between one-dot core size $NM^2$ and training dataset size $|\mathcal{D}^\prime|=(f+1)|\mathcal{D}|=4,375$ is noteworthy.
The Coordinator layer is effective for smaller bond dimensions ($M=7,14$) but becomes redundant at $M=14$ and counterproductive at $M=28$, likely because the tensor train layer itself achieves sufficient expressivity at larger $M$ values.
This suggests that for larger systems with constrained bond dimension $M$, the Coordinator layer may enhance performance, or pretraining with small $M$ and the Coordinator layer could improve results.
Subsequent results employed $M=14$, $N=21$, with the optimized Coordinator layer.
The training trace is shown in Fig.~\ref{fig:nnmpo-trace}, 
\magenta{which required approximately one hour of total wall clock time using an NVIDIA A100 GPU with 80GB VRAM.}
For the basis optimization, we divided the 625 training points into five mini-batches, each containing 125 points.
For the first 1,000 epochs, to make a good initial guess, one-dot sweeping and basis optimization by Adam were performed iteratively and confirmed the loss function was roughly converged.
From 1,000 epochs to 16,000 epochs, to increase a representational power, two-dot sweeping was executed every 500 epochs and switched to one-dot sweeping once the bond dimension reached 14.
Finally, we fixed the basis and refined TT by one-dot sweeping using the CG method until it tightly converged.
The final RMSE of the energy for the validation set reached less than 1.0 meV.
The scatter plot between the predicted energy by NN-MPO with $N=21$ and $M=14$ and test energy is shown in Fig.~\ref{fig:nnmpo-scatter}
and the mean absolute error (MAE) was 3.03 \kayser.
Fig.~\ref{fig:nnmpo-coordinator} shows the matrix elements of the coordinator $U$ of the NN-MPO after training, which indicates that initial normal modes are suitable for the description of the PES in a low-rank tensor network because the coordinator matrix is almost the identity matrix.
If we examine the coordinator matrix further, we see that the $x_3$ and $x_4$ are slightly mixed.
This should be because both have the same symmetry, $A_1$ irreducible representation, and close harmonic frequencies around 1,500 to 1,800 \kayser.
The displacement vector of $q_4$ (C=O stretching) is a little bit different from $x_4$ in terms of the $\angle \text{H}_1\text{OH}_2$ angle.
The $x_1$, unique out-of-plane mode, is the most separated from the others.
In the present results, significant rotations of the coordinate were not observed. 
However, we initially had anticipated that similar to how diagonalizing the Hessian matrix provides the coordinates that yield the lowest-rank, TT-rank 2, structure in the HO approximation, \magenta{which is described in Appendix~\ref{sec:ho-bond-dim}}. 
\magenta{
Previous research has demonstrated that optimizing linear coordinates for realistic potentials yields advantageous coordinates for subsequent quantum mechanical calculations
\cite{thompson1982truhlar,yagi2012hirata,jacob2009reiher,hino2022kurashige}. 
Similarly, training the Coordinator layer can produce an MPO representation with a small bond dimension, which is expected to reduce the bond dimension of the MPS wavefunction upon which the MPO repeatedly acts, particularly during the spanning of Krylov subspaces in subsequent quantum mechanical calculations. 
}
\magenta{
NN-MPO was trained with force-augmented data in this work; however, when force data is unavailable as is often the case with higher-level electronic structure calculations where analytical force computation is challenging, additional energy data points become necessary to compensate. For instance, in the case where $N=21$ and $M=14$, training with energy data alone leads to overfitting because the size of a single TT core ($NM^2$) substantially exceeds the number of available training data points (625).
For comparative analysis, we fitted the same dataset using a MLP with SiLU activation function, 6 hidden layers with 48 neurons per layer and residual connections. This model underwent training for 100,000 iterations with identical loss function and Adam optimizer configurations. The training process was completed in approximately 30 minutes with a single NVIDIA A100 GPU. The MLP achieved a MAE of 3.61 \kayser, which demonstrates performance comparable to our NN-MPO approach.
}
For comprehensive implementation details, including hyperparameter specifications, we provide our Python NN-MPO implementation at \url{https://github.com/KenHino/Pompon}. This implementation accepts diverse datasets containing atomic positions, energies, and forces (optional) in NumPy array format. The trained MPO can be exported to Numpy array and HDF5 format, ensuring compatibility with the \texttt{ITensors.jl} \magenta{or \texttt{ITensorMPS.jl}} frameworks for subsequent quantum mechanical analyses, including eigenvalue calculations via DMRG and quantum dynamics simulations using the time-dependent variational principle (TDVP) methodology \cite{haegeman2016verstraete}.

\subsection{DMRG calculations}
\begin{figure}
  \centering
  \includegraphics[width=0.5\columnwidth]{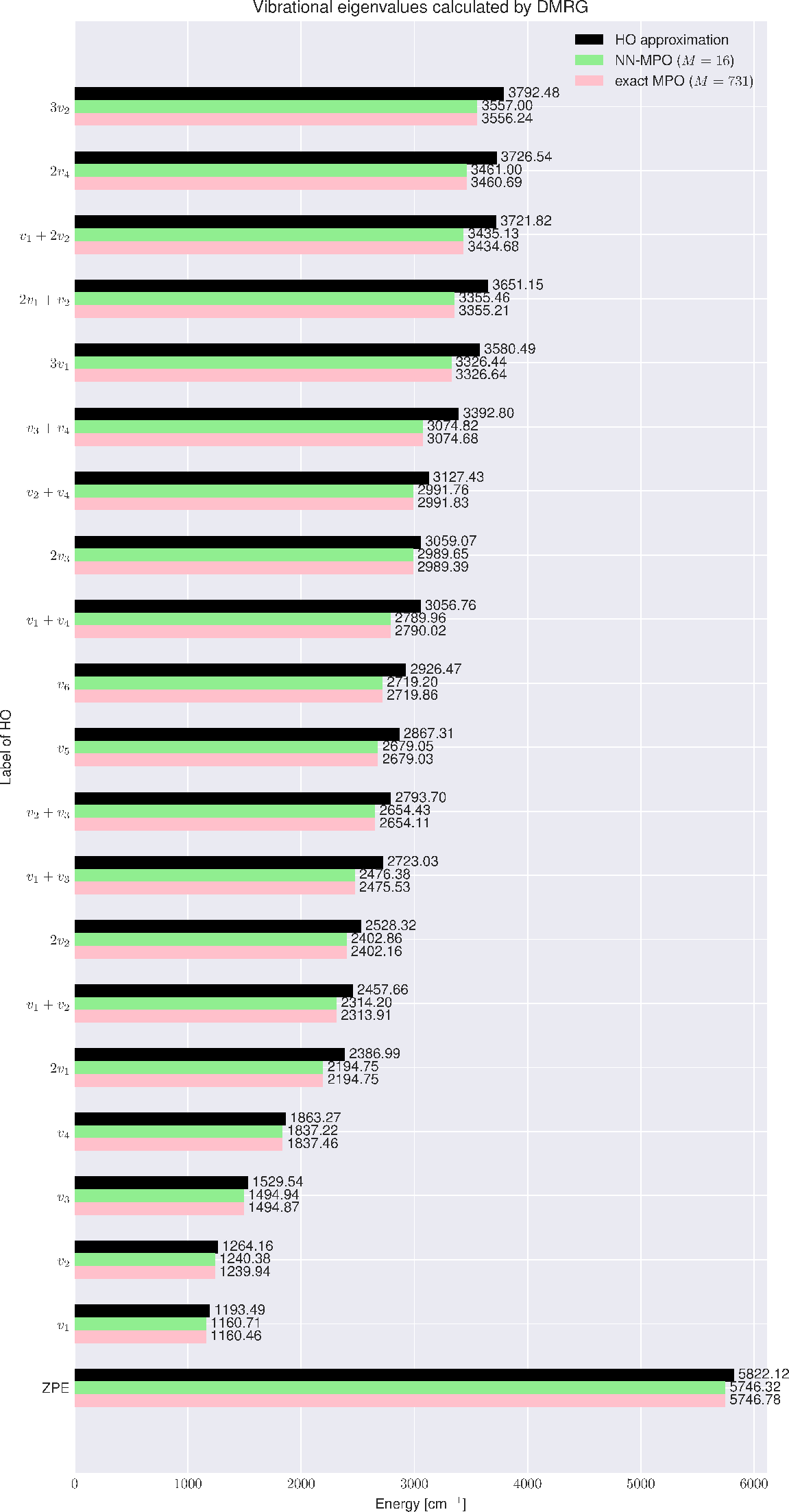}
  \caption{
  \label{fig:nnmpo-eigval-hist}
    Vibrational eigenvalues of the \form molecule
    calculated by HO approximation and 
    DMRG using NN-MPO and exact MPO.
    The HO approximation represents eigenvalues obtained from the harmonic oscillator Hamiltonian, which is constructed using the Hessian matrix evaluated at the equilibrium geometry.
  }
\end{figure}

After NN-MPO had been trained, we solved the Schr\"odinger equation, Eq.~(\ref{eq:hamiltonian-mw-hidden-nnmpo}), to obtain the first 21 vibrational levels of the \form molecule.
We chose wavefunction basis $\Ket{\sigma_i}$ as HO discrete variable representation (DVR) \cite{colbert1992miller,beck2000meyer}, which is derived from the diagonalization of the representation matrix between HO eigenfunctions and position operator, and enables us to approximate the integral between position-dependent function and DVR basis by evaluating the function value at DVR grid points.
We used our DVR basis implementation available at \url{https://github.com/KenHino/Discvar}.
DVR grid basis $\Ket{n}$ has been facilitated by MPS and MPO framework \cite{hino2024kurashige}.
Both the kinetic operator and NN-MPO were converted to MPO, and phonon DMRG calculation was performed by using \texttt{ITensors.jl} library version 0.6 available at \url{https://github.com/ITensor/ITensors.jl} \cite{fishman2022stoudenmire}.
Kinetic MPO is given by $M=2$;
\begin{equation}
  \label{eq:kinetic-mpo}
  -\frac{\hbar^2}{2}
  \begin{bmatrix}
    \hat{p}_1^2 & 1
    \end{bmatrix}
    \begin{bmatrix}
    1 & 0 \\
    \hat{p}_2^2 & 1
    \end{bmatrix}
    \cdots
    \begin{bmatrix}
    1 & 0 \\
    \hat{p}_{f-1}^2 & 1
    \end{bmatrix}
    \begin{bmatrix}
    1 \\
    \hat{p}_{f}^2
    \end{bmatrix}
\end{equation}
where $\hat{p}_i$ is the momentum operator of $i$-th basis.
We set the number of wavefunction basis $d=9$ and the maximum bond dimension of MPS $m=100$. 
The vibrational levels calculated by NN-MPO are shown in Fig~\ref{fig:nnmpo-eigval-hist} and compared with HO approximation and the full-dimensional mesh grid potential (exact MPO), which requires DFT calculations for all possible $d^6=531,441$ DVR grid points and reaches its bond dimension to $M=d^3=729$ without any compression.
\magenta{The MAE between NN-MPO and exact MPO for up to twenty excited vibrational levels was 0.325 \kayser.
}
This result suggests that by utilizing datasets derived from higher-level electronic structure calculations, we could reproduce experimental spectroscopic measurements with exceptional accuracy.
%

\section{Conclusions}
We have introduced the neural network matrix product operator (NN-MPO), which can be trained with energy and force at randomly sampled points and is compatible with multi-dimensional integral.
We have successfully applied a six-dimensional system to demonstrate that the NN-MPO can predict DFT energies with MAE of 3.03 \kayser, where the energies reach up to 17,000 \kayser, using only 625 training data points.
In addition, we have introduced the coordinator $U$ as a trainable parameter, orthogonal linear transformation matrix from the input mass-weighted coordinates $\mathbf{x}$, which would yield a suitable coordinate for quantum mechanics calculations because the kinetic energy operator is kept the simple form and the latent coordinate leads to a lower-rank representation.
\magenta{
Utilizing the trained NN-MPO, we conducted phonon DMRG calculations and evaluated the MAE for up to twenty vibrational excited states between the NN-MPO and exact MPO representations, achieving spectroscopic accuracy of 0.325 \kayser.
}
%
However, several challenges remain.
For instance, the symmetry of the molecule should be considered, which not only enhances the sampling efficiency but also avoids nonphysical predictions such as artificially lifted degeneracies in vibrational spectra during quantum static calculations and nonphysical branching ratios to final states in quantum dynamical simulations.
The proper coordinate can introduce translational and rotational invariance, although sophisticated internal coordinates, such as polyspherical coordinates, tend to complicate the kinetic energy operator (KEO).
For example, incorporating Fourier basis functions into the NN-MPO framework for angular degrees of freedom in polar coordinates can partially address rotational invariance requirements.
In particular, extending full-dimensional simulations to flexible molecular systems, such as those containing methyl groups or water clusters, these techniques are essential.
\magenta{Some techniques have been proposed to address the algebraic complexity of KEO in curvilinear coordinates}
\cite{lauvergnat2002tnum,nauts1985andchapuisat,bader2023christiansen,grinfeld2013grinfeld}.
The adoption of the permutational invariance is not straightforward.
In the context of SOP formalism, a method has been proposed to restore symmetries using symmetric operators $\mathcal{R}_n$ \cite{schroder2020schroder}.
This approach represents the prediction model as an average over symmetry operations: $V_{\text{sym}}(Q_1, Q_2, \cdots, Q_f) = \frac{1}{N_{\text{sym}}} \sum_n^{N_{\text{sym}}} \mathcal{R}_nV(Q_1, Q_2, \cdots Q_f)$.
This symmetrization technique can be readily extended to the NN-MPO framework.
The permutationally invariant polynomial (PIP) 
\cite{braams2009bowman,jiang2013guo,jiang2016guo,koner2020meuwly}
also may give us a hint on how to address this issue.
The quantum number conservation approach, which is used for electronic MPS wavefunction, may enhance the capability of the NN-MPO.
Furthermore, the scalability involving the number of parameters, training data points, and optimization convergence has not yet been explored deeply.
Ideally, NN-MPO can express PES with a scalable number of parameters if the system involves short-range interactions, such as hydrogen bonds or Van der Waals interactions, or exhibits a one-dimensional structure, like polyenes.
The key to extending systems with dozens of atoms may lie in employing localized coordinates,
which effectively reduce the entanglement in both wavefunctions and PESs.
Although we acknowledge that extending to full-dimensional simulations of systems containing hundreds or even dozens of atoms remains computationally prohibitive, for practical applications, we can effectively address this limitation by focusing on active modes that dominantly contribute to the physical quantities of interest.
By strategically selecting modes localized to specific atomic groups, we can substantially mitigate entanglement across the system and constrain the bond dimension to manageable values, thereby extending the applicability of our approach to larger molecular systems.
The Coordinator layer we introduced has the potential to generate localized coordinates that are particularly suitable for one-dimensional TNs during the training process.
The CG method is a robust approach to solving quadratic optimization problems; 
however, it becomes challenging when the number of data points $|\mathcal{D}|$ is enormous, particularly as the system size increases.
$Novikov\;et\;al$ have proposed a different optimization scheme for TT \cite{novikov2017oseledets} facilitated by the Riemannian optimization, which is oriented to mini-batch optimization.
\magenta{
While controlling the bond dimension $M$ inherently serves as a regularization technique, additional regularization methods such as L2 regularization and dropout can potentially enhance the performance of the NN-MPO. Furthermore, implementing appropriate initialization strategies and entropy regularization for basis functions may contribute to reducing the required number of basis functions $N$.
}
The NN-MPO can be applied not only to MPS wavefunctions but also to a linear combination of state vectors and mean-field wavefunctions. 
In contrast, the SOP is typically employed for a wide range of tensor networks, 
including tree tensor networks like the multi-layer MCTDH \cite{wang2003thoss}.
Despite these issues, our method offers a promising alternative for quantum many-body simulations. It bridges the gap between traditional SOP representations and modern machine learning techniques and allows us to approach the Born-Oppenheimer approximation limit accurately. 
Moreover, multi-dimensional integrals arise in various contexts, such as evaluating expectation values, variational optimization, marginalizing degrees of freedom, and calculating free energy using configuration integrals. 
By harnessing the capabilities of the NN-MPO for these applications, we can position it as a versatile instrument not only within quantum mechanics but across a broader spectrum of scientific disciplines.

\begin{acknowledgments}
This work was supported by JSPS KAKENHI (JP23KJ1334, JP23H01921),
JST-FOREST Program (JPMJFR221R), JST-CREST Program (JPMJCR23I6), and
MEXT Q-LEAP Program (JPMXS0120319794).
\end{acknowledgments}

\appendix
\magenta{
\section{Bond dimension of harmonic oscillator with and without the coordinate transformation}
}
\label{sec:ho-bond-dim}
To elucidate the impact of coordinate transformation, we examine the harmonic potential, which provides the most intuitive and quantitatively tractable example. 
For a 6-dimensional second-order polynomial potential, the mathematical expression is given by
$$
V(x_1, x_2, \cdots, x_6)
= 
\begin{pmatrix}
 x_1 & x_2 & \cdots & x_6
\end{pmatrix}
\begin{pmatrix}
  H_{11} & \frac{H_{12}}{2} & \cdots & \frac{H_{16}}{2} \\
  \frac{H_{12}}{2} & H_{22} & \cdots & \frac{H_{26}}{2} \\
  \vdots & \vdots & \ddots & \vdots \\
  \frac{H_{16}}{2} & \frac{H_{26}}{2} & \cdots & H_{66}
\end{pmatrix}
\begin{pmatrix}
  x_1 \\
  x_2 \\
  \vdots \\
  x_6
\end{pmatrix},
$$
which requires $M=5$ and $N=3$ structure in MPO representation:
$$
\begin{aligned}
\begin{pmatrix}
  x_1^2 & x_1 & 1
\end{pmatrix}
\begin{pmatrix}
  0 & 0 & 0 & H_{11} \\
  0 & 1 & 0 & H_{12} x_2 \\
  x_2 & 0 & 1 & H_{22} x_2^2
\end{pmatrix}
\begin{pmatrix}
  0 & 0 & 1 & 0 & H_{23} x_3 \\
  1 & 0 & 0 & 0 & H_{13} x_3 \\
  0 & x_3 & 0 & 1 & H_{33} x_3^2 \\
  0 & 0 & 0 & 0 & 1
\end{pmatrix}
\\
\begin{pmatrix}
  H_{14} x_4 & H_{16} & H_{15} & 0 \\
  H_{34} x_4 & H_{36} & H_{35} & 0 \\
  H_{24} x_4 & H_{26} & H_{25} & 0 \\
  H_{44} x_4^2 & H_{46} x_4 & H_{45} x_4 & 1 \\
  1 & 0 & 0 & 0
\end{pmatrix}
\begin{pmatrix}
  1 & 0 & 0 \\
  0 & 1 & 0 \\
  x_5 & 0 & 0 \\
  H_{55} x_5^2 & H_{56} x_5 & 1
\end{pmatrix}
\begin{pmatrix}
  1 \\
  x_6 \\
  H_{66} x_6^2
\end{pmatrix}
\end{aligned}
$$
On the other hand, if the coordinator layer captures the appropriate coordinate $\mathbf{q} = \mathbf{x} U$, the normal coordinate in this case, the potential is given by
$$
V(q_1, q_2, \cdots, q_6)
= 
\begin{pmatrix}
  q_1 & q_2 & \cdots & q_6
\end{pmatrix}
\begin{pmatrix}
\Lambda_1 & 0 & \cdots & 0 \\
0 & \Lambda_2 & \cdots & 0 \\
\vdots & \vdots & \ddots & \vdots \\
0 & 0 & \cdots & \Lambda_6
\end{pmatrix}
\begin{pmatrix}
  q_1 \\
  q_2 \\
  \vdots \\
  q_6
\end{pmatrix}
$$
where $H = U \Lambda U^{\top}$.  
This requires $M=2$ and $N=2$ structure in MPO representation:
$$
\begin{pmatrix}
\Lambda_1 q_1^2 & 1
\end{pmatrix}
\begin{pmatrix}
1 & 0 \\
\Lambda_2 q_2^2 & 1
\end{pmatrix}
\begin{pmatrix}
1 & 0 \\
\Lambda_3 q_3^2 & 1
\end{pmatrix}
\begin{pmatrix}
1 & 0 \\
\Lambda_4 q_4^2 & 1
\end{pmatrix}
\begin{pmatrix}
1 & 0 \\
\Lambda_5 q_5^2 & 1
\end{pmatrix}
\begin{pmatrix}
1 \\
\Lambda_6 q_6^2
\end{pmatrix}
$$
While realistic molecular systems exhibit significantly more complexity than this simplified example, the coordinator layer is still expected to facilitate a moderate reduction in bond dimension through effective coordinate transformation.

\section{Conjugate gradient method for TT core optimization}
\label{sec:cg}
Sweeping optimization with tight convergence criterion is the most time-consuming part of the NN-MPO training process. In this process, we aim to minimize the function
\begin{equation}
  f\left(\boldsymbol{x}\right) = \frac{1}{2}\boldsymbol{x}^\top A \boldsymbol{x} - \boldsymbol{x}^\top b
  \tag{\ref{eq:quadratic-form}}
\end{equation}
where $\boldsymbol{x}$ is the vectorized $B\substack{\rho_i\rho_{i+1}\\\beta_{i-1}\beta_{i+1}}$,
the Hessian matrix is 
$A=\sum_p\left(
  \varPhi\substack{p\\\rho_i^\prime\rho_{i+1}^\prime\\\beta_{i-1}^\prime\beta_{i+1}^\prime}
\right)^\top \BlockAll$ 
and $b=\sum_p\left(\BlockAll\right)^\top y_p$. For this optimization problem, the conjugate gradient (CG) method provides a robust solution.
We can avoid explicitly storing the Hessian matrix $A\in\mathbb{R}^{M^2N^2\times M^2N^2}$, as it can be constructed on-the-fly from $\BlockL$, $\BlockR$, $\varphi^{p}_{\rho_i}$ and $\varphi^{p}_{\rho_{i+1}}$.
Consequently, the computational complexity for evaluating $A\boldsymbol{x}$ is $\mathcal{O}(M^2N^2|\mathcal{D}^\prime|)$. 
Assuming that the total number of iterations $k_{\text{max}}$ scales as the square root of the vector size, the computational complexity of sweeping optimization, which involves repeating TT core optimization via the conjugate gradient method for $\mathcal{O}(MN)$ iterations $\mathcal{O}(f)$ times, becomes $\mathcal{O}(fM^3N^3|\mathcal{D}^\prime|) = \mathcal{O}(f^2M^3N^3|\mathcal{D}|)\;(\because |\mathcal{D}^\prime| = (f+1)|\mathcal{D}|)$ \cite{stoudenmire2016schwab}.
In contrast, conventional kernel trick scales at least as $\mathcal{O}(|\mathcal{D}^\prime|^2) = \mathcal{O}(f^2|\mathcal{D}|^2)$ 
without employing sophisticated techniques \cite{cesa-bianchi2015shamir}.
Both NN-MPO sweeping optimization and kernel trick scales quadratically with system size $\mathcal{O}(f^2)$ when training with force data, but kernel trick also scales quadratically with the number of training data points $\mathcal{O}(|\mathcal{D}|^2)$, whereas NN-MPO sweeping optimization scales linearly $\mathcal{O}(|\mathcal{D}|)$ which can be a key advantage to adopt NN-MPO for large-scale datasets.
It is worth noting that interatomic neural network potentials achieve scalability with respect to both training data points $|\mathcal{D}|$ and system size $f$ by decomposing forces to individual atoms.
The complete algorithm is presented in Algorithm~\ref{alg:cg}.
\magenta{We note that L2 regularization can be incorporated by replacing $A$ with $A+\lambda I$, which may be particularly appropriate when the number of parameters in a TT core exceeds the number of training data points.}

\begin{algorithm}[H] 
  \caption{Conjugate gradient method for TT core optimization} 
  \label{alg:cg}
  \begin{algorithmic}
    \State $\BlockL \in \mathbb{R}^{|\mathcal{D^\prime}|\times M}$, 
    $\BlockR\in \mathbb{R}^{|\mathcal{D^\prime}|\times M}$,
    $\varphi^{p}_{\rho_i} \in \mathbb{R}^{|\mathcal{D}^\prime|\times N}$, 
    $\varphi^{p}_{\rho_{i+1}} \in \mathbb{R}^{|\mathcal{D}^\prime|\times N}$,
    $y_p \in \mathbb{R}^{|\mathcal{D}^\prime|}$, 
    $B, b, r_k, p_k \in \mathbb{R}^{M \times N \times N \times M}$
  \end{algorithmic}
  \begin{algorithmic}[1]
    \Function {GetAx}
    {$\BlockL, \BlockR, \varphi^{p}_{\rho_i}, \varphi^{p}_{\rho_{i+1}}, 
      x^{\rho_i\rho_{i+1}}_{\beta_{i-1}\beta_{i+1}}$}
      \State \Return $
      \sum_{p,\rho_i,\rho_{i+1},\beta_{i-1},\beta_{i+1}}
      \newline 
      \varPhi^{[:i-1]}_{p\beta_{i-1}^\prime} \varPhi^{[i+2:]}_{p\beta_{i+1}^\prime} 
      \varphi^{p}_{\rho_i^\prime} \varphi^{p}_{\rho_{i+1}^\prime} 
      \BlockL\BlockR
      \varphi^{p}_{\rho_i} \varphi^{p}_{\rho_{i+1}} 
      x^{\rho_i\rho_{i+1}}_{\beta_{i-1}\beta_{i+1}}
      $
    \EndFunction
  \end{algorithmic}
  \begin{algorithmic}[1]
    \Function {ConjugateGradient}
      {$\BlockL$, 
      $\BlockR$,
      $\varphi^{p}_{\rho_i}$, 
      $\varphi^{p}_{\rho_{i+1}}$,
      $y_p$, 
      $B^{\rho_i\rho_{i+1}}_{\beta_{i-1}\beta_{i+1}}$, 
      $k_{\text{max}}$, 
      $\epsilon$}
      \State $B_0 \gets \TwoDot$
      \State $b \gets \sum_p \BlockL\BlockR
      \varphi^{p}_{\rho_i}\varphi^{p}_{\rho_{i+1}}y_p$
      \State $r_0 \gets b - \text{GetAx}\left(
      \BlockL, \BlockR,
      \varphi^{p}_{\rho_i}, \varphi^{p}_{\rho_{i+1}}, B_0
      \right)$
      \State $p_0 \gets r_0$
      \State $k \gets 0$
      \While{$k < k_{\text{max}}$}
        \State $Ap_k \gets \text{GetAx}\left(
        \BlockL, \BlockR,
        \varphi^{p}_{\rho_i}, \varphi^{p}_{\rho_{i+1}}, p_k
        \right)
        $
        \State $p_k^\top Ap_k \gets \sum_{\substack{\rho_i\rho_{i+1}\\\beta_{i-1}\beta_{i+1}}} 
        \left(p_k\right)\substack{\rho_i\rho_{i+1}\\\beta_{i-1}\beta_{i+1}}
        \left(Ap_k\right)\substack{\rho_i\rho_{i+1}\\\beta_{i-1}\beta_{i+1}}$
        \State $\alpha_k \gets \frac{\|r_k\|^2}{p_k^\top Ap_k}$
        \State $B_{k+1} \gets B_{k} + \alpha_k p_k$
        \State $r_{k+1} \gets r_{k} - \alpha_k A p_k$
        \If{$\|r_{k+1}\| < \epsilon$}
          \State \textbf{break}
        \EndIf
        \State $\beta_k \gets \frac{\|r_{k+1}\|^2}{\|r_k\|^2}$
        \State $p_{k+1} \gets r_k + \beta_k p_k$
        \State $k \gets k + 1$
      \EndWhile
      \State \Return $B_{k+1}$
    \EndFunction
  \end{algorithmic}
\end{algorithm}

\bibliography{ms}

\end{document}